\documentclass{article}

\usepackage{microtype}
\usepackage{graphicx}
\usepackage{hyperref}
\usepackage{bm}
\usepackage{subfigure}
\usepackage{amsmath}
\usepackage{multirow}
\usepackage{array}
\usepackage{booktabs}
\usepackage{pifont}
\usepackage{colortbl}
\usepackage{booktabs}
\usepackage{afterpage}
\usepackage{amssymb} 
\usepackage{booktabs} 
\usepackage{lineno}
\usepackage{hyperref}

\usepackage{hyperref}

\usepackage[accepted]{icml2024}

\usepackage{amsmath}
\usepackage{amssymb}
\usepackage{mathtools}
\usepackage{amsthm}

\usepackage[capitalize,noabbrev]{cleveref}

\theoremstyle{plain}

\theoremstyle{definition}

\theoremstyle{remark}
\usepackage{bm}

\usepackage[textsize=tiny]{todonotes}

\icmltitlerunning{SynArtifact: Classifying and Alleviating Artifacts in Synthetic Images via Vision-Language Model}

\begin{document}

\hypersetup{
}
\newcommand{\red}[1]{\textcolor{red}{#1}}

\twocolumn[
\icmltitle{SynArtifact: Classifying and Alleviating Artifacts \\ in Synthetic Images via Vision-Language Model}



\icmlsetsymbol{equal}{*}

\begin{icmlauthorlist}
\icmlauthor{Bin Cao}{casia,ucas,baai}
\icmlauthor{Jianhao Yuan}{Oxford}
\icmlauthor{Yexin Liu}{baai}
\icmlauthor{Jian Li}{Tencent}
\icmlauthor{Shuyang Sun}{Oxford}
\icmlauthor{Jing Liu}{casia,ucas}
\icmlauthor{Bo Zhao}{baai}
\end{icmlauthorlist}

\icmlaffiliation{casia}{Institute of Automation, Chinese Academy of Sciences}
\icmlaffiliation{ucas}{School of Artificial Intelligence, University of Chinese Academy of Sciences}
\icmlaffiliation{baai}{Beijing Academy of Artificial Intelligence}
\icmlaffiliation{Oxford}{University of Oxford}
\icmlaffiliation{Tencent}{YouTu Lab, Tencent}

\icmlcorrespondingauthor{Bo Zhao}{bozhaonanjing@gmail.com}

\icmlkeywords{Machine Learning, ICML}

\vskip 0.3in
]



\printAffiliationsAndNotice{}  

\begin{abstract}
    In the rapidly evolving area of image synthesis, a serious challenge is the presence of complex artifacts that compromise perceptual realism of synthetic images. To alleviate artifacts and improve quality of synthetic images, we fine-tune Vision-Language Model (VLM) as artifact classifier to automatically identify and classify a wide range of artifacts and provide supervision for further optimizing generative models. Specifically, we develop a comprehensive artifact taxonomy and construct a dataset of synthetic images with artifact annotations for fine-tuning VLM, named SynArtifact-1K. The fine-tuned VLM exhibits superior ability of identifying artifacts and outperforms the baseline by 25.66\%. To our knowledge, this is the first time such end-to-end artifact classification task and solution have been proposed. Finally, we leverage the output of VLM as feedback to refine the generative model for alleviating artifacts. Visualization results and user study demonstrate that the quality of images synthesized by the refined diffusion model has been obviously improved. The dataset is available at \url{https://github.com/BBBiiinnn/SynArtifact}
\end{abstract}

\section{Introduction}
\label{sec:intro}
The field of image synthesis has witnessed remarkable advancements, primarily driven by the development of generative models \cite{zhang2023adding,ho2020denoising,ramesh2022hierarchical,saharia2205photorealistic}. These models have found widespread applications across various domains, including art creation, medical imaging, and autonomous driving. Despite this progress and applications, the quality of synthetic images still does not align with human preference perfectly. 

Some related works \cite{xu2023imagereward,huang2023t2i} employ reward-driven methods to optimize generative models for improving the quality of synthetic images. However, most existing methods mainly depend on some single-score evaluation metrics of synthetic images. Utilizing single-score metric for optimizing generative models exhibits the inherent limitation, lacking the ability to reflect the diversity and complexity of artifacts directly, a significant issue that undermines the quality and perceptual realism of synthetic images. For example, a human annotator probably assigns the same score to a synthetic image with different kinds of artifacts, for example an image featuring distorted fingers (\textbf{Distortion}) while another image featuring six fingers (\textbf{Duplication}). Consequently, such less-informative annotations lead to ineffective attempt of automatically alleviating distortion and duplication in synthetic images. 

The main challenge of artifact classification is the diversity and complexity of artifacts. To address this, we first develop a comprehensive taxonomy of common artifacts found in images synthesized by various generative models and construct a dataset with human-annotated artifact categories, captions and coordinates of artifacts, named SynArtifact-1K. The dataset is utilized for fine-tuning Vision-Language Model (VLM) to classify a wide range of artifacts. Experimental results show that the VLM fine-tuned on SynArtifact-1K surpasses the baseline by 25.66\% of classification accuracy and 29.01\% of F1 score. To alleviate artifacts, we improve the generative model by leveraging output of artifact classifier as the AI feedback, namely, Reinforcement Learning from AI Feedback (RLAIF) \cite{bai2022constitutional}. Extensive results verify that the fine-tuned VLM can serve as the artifact classifier providing guidance for optimizing diffusion models. Our main contributions can be summarized as follows:

\begin{figure*}[t]
\centering
    \includegraphics[width=1.0\textwidth]{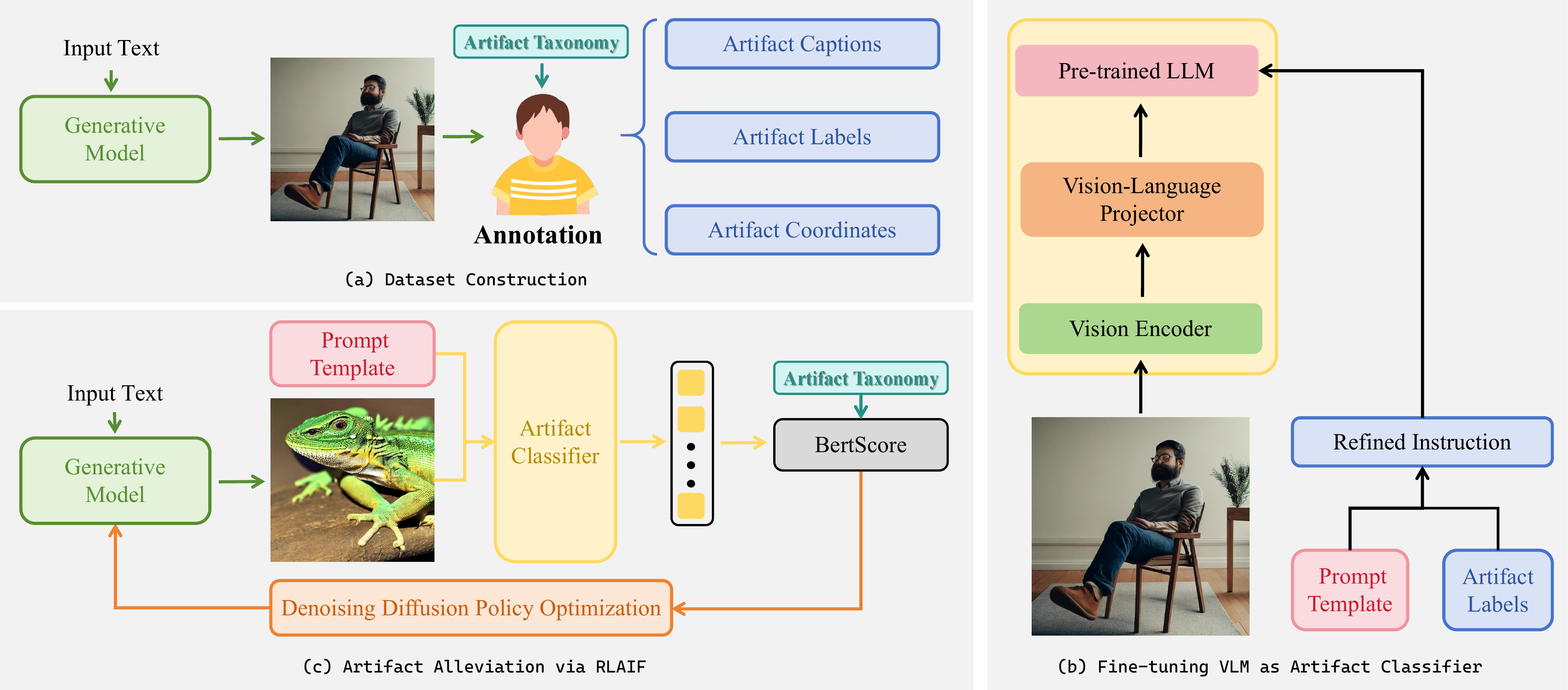}
\caption{\textbf{The Proposed Method.} The entire pipeline consists of three components: dataset construction (see \cref{sec:artifacts taxonomy and dataset construction}), fine-tuning VLM as artifact classifier (see \cref{sec:Automatic Artifact Detection with Vision-Language Model}) and artifact alleviation via RLAIF (see \cref{sec:Detection4RL}).}
\label{fig:pipeline}
\end{figure*}

\begin{itemize}
    \item We design a comprehensive artifact taxonomy for synthetic images including 13 kinds of artifacts, and propose the first image-with-artifact dataset, named SynArtifact-1K. The dataset contains 1.3k annotated images generated by various generative models, and each image is annotated with categories, captions and coordinates of artifacts.
    \item We demonstrate that VLM fine-tuned on SynArtifact-1K outperforms the baseline by 25.66\% of classification accuracy and 29.01\% of F1 score, showing the ability to identify artifacts and provide supervision for optimizing diffusion models.  
    \item We exploit output of the fine-tuned VLM, namely, artifact classifier, as the supervision to refine the diffusion model in the manner of RLAIF. Extensive experiments demonstrate the refined diffusion model produces higher-quality images with less artifacts.
\end{itemize}

\section{Related Work}
\label{sec:related_work}

\subsection{Synthetic Image Assessment}
Inception Score (IS) \cite{salimans2016improved} and Fréchet Inception Distance (FID) \cite{heusel2017gans} are widely-used automatic evaluation metrics for evaluating the performance of generative models. Both can only measure the feature distance between real and synthetic images.  
However, several related works reveal that IS and FID do not align with human preferences \textcolor{blue}{\cite{wu2023human,kirstain2023pick}}. AGIQA-1K \cite{zhang2023perceptual} constructs a text-to-image dataset with human preference and conducts synthetic image quality evaluation through MOS (Mean Opinion Score). Pick-a-Pic \cite{kirstain2023pick} trains a CLIP-based score function \cite{radford2021learning} for predicting human preferences. HPSv2 \cite{wu2023human} utilizes ChatGPT to rewrite prompts from DiffusionDB \cite{wangDiffusionDBLargescalePrompt2022} and tunes CLIP through minimizing KL-divergence between the CLIP Score and human preference. ImageReward \cite{xu2023imagereward} employ BLIP \cite{li2022blip} as backbone and MLP head to predict comparison of synthetic images. However, these methods for assessing synthetic images overlook artifacts and primarily focus on scoring or ranking. 

\begin{figure*}[t]
\centering
    \includegraphics[width=\textwidth]{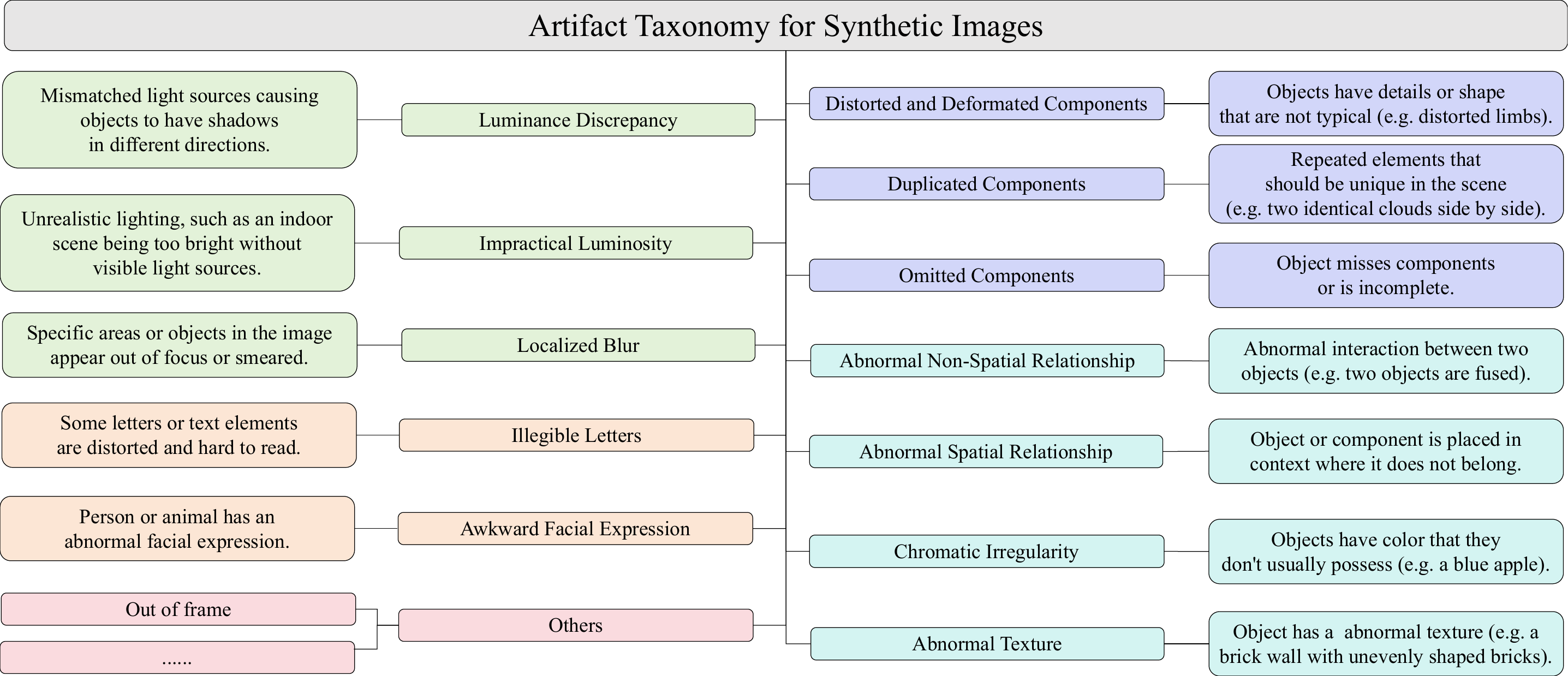}
\caption{\textbf{Artifact Taxonomy.} The entire artifact taxonomy contains 13 kinds of artifacts. Each kind of artifact is accompanied with explanation.}
\label{fig:Artifact Taxonomy}
\end{figure*}

Another line of works focus on the text-to-image alignment evaluation. CLIP Score \cite{hessel2021clipscore} computes cosine similarity between image embedding and text embedding as the metric for text-to-image alignment. T2I-CompBench \cite{huang2023t2i} is a comprehensive benchmark for text-to-image generation, evaluating text-to-image alignment from 3 categories including attribute binding, object relationship and complex compositions. GENEVAL \cite{ghosh2023geneval} proposes an object-focused framework for evaluating count, color and position of objects, leveraging object detection models. Previous object-level text-to-image alignment evaluation methods \cite{ghosh2023geneval} \cite{Cho2023VPT2I} utilize detection model or segmentation model. In our work, we attempt to construct an end-to-end framework for artifact classification without any object detection models.

\subsection{Assessment with Vision-Language Model}
Vision-Language Models have achieved successful progress in recent years \cite{chen2023minigptv2,zhu2023minigpt,liu2023llava,liu2023improvedllava,chen2023shikra}, especially integrating with LLMs (Large Language Models) \cite{zheng2023judging,touvron2023llama}. X-IQE \cite{chen2023x} utilizes MiniGPT-4 \cite{zhu2023minigpt} and designs a hierarchical CoT (Chain of Thought) \cite{wei2022chain} to generate text for assessment. The entire CoT includes fidelity evaluation, alignment evaluation, and aesthetic evaluation. Each evaluation task consists of an image description, task-specific analysis, and scoring. TIFA \cite{hu2023tifa} uses GPT-3 \cite{mann2020language} to generate question-answer pairs filtered by a QA model. Then, the VQA accuracy is the metric for faithfulness between the synthetic image and text prompt. VPEval \cite{Cho2023VPT2I} defines five kinds of image generation skills, including object, count, spatial scale, and text rendering. Subsequently, it designs a series of specialized modules for assessment. DepictQA \cite{you2023depicting} breaks through the constraints of score-based methods, leveraging Multi-modal Language Models for image quality evaluation. However, DepictQA \cite{you2023depicting} utilizes a reference image for comparison. In our work, we aim to construct a artifact classification model without reference.

\subsection{Reinforcement Learning from AI Feedback}
Reinforcement Learning from Human Feedback (RLHF) \cite{ouyang2022training}, which fine-tunes language models towards human preference, is initially adopted in natural language processing (NLP). However, this method needs massive expert-annotated model outputs, which remains time-consuming and challenging. Reinforcement Learning from AI Feedback (RLAIF) is first introduced in \cite{bai2022constitutional}, which provides a promising alternative. Lee et al. \cite{lee2023rlaif} demonstrates RLAIF can achieve comparable performance to RLHF on summarization, helpful dialogue generation, and harmless dialogue generation through experiments. In this paper, we employ RLAIF to guide the optimization of diffusion model online.

\section{Methodology}
\label{sec:methodology}

\subsection{Overview}

\cref{fig:pipeline} gives an overview of our approach. To classify and alleviate artifacts, we first employ diffusion model to generate synthetic images and annotate images under the guidance of a comprehensive artifact taxonomy resulting in artifact labels, captions and coordinates. Next, we convert artifact labels into visual instruction utilizing the prompt template. Then, we employ visual instructions to fine-tune VLM as artifact classifier. Finally, leveraging output of fine-tuned VLM, namely, artifact classifier, as AI feedback, \emph{BertScore} between output of artifact classifier and each kind of artifact is calculated as artifact classification reward. Diffusion model is optimized by maximizing artifact classification reward to alleviate artifacts through Denoising Diffusion Policy Optimization (DDPO) \cite{black2023training}.

\subsection{Dataset Construction}
\label{sec:artifacts taxonomy and dataset construction}
\subsubsection{Artifact Taxonomy}
\label{sec:artifacts taxonomy}
While substantial advancements have been achieved in generative models, the synthetic images still struggle to align with human preferences. Moreover, the artifacts, which compromise quality of synthetic images, remain common in synthetic images. Therefore, as the first step of alleviating artifacts, we analyze and categorize various types of artifacts presented in synthetic images. 

We construct a comprehensive taxonomy of artifacts, as shown in \cref{fig:Artifact Taxonomy}. In the coarse-grained classification, artifacts can be categorized into four groups w.r.t. \emph{object-aware}, \emph{object-agnostic}, \emph{lighting}, and \emph{others}. The popular object-aware related artifacts include illegible letters which are hard to recognize, and awkward facial expressions of humans and animals. Object-agnostic related artifacts contain two sub-categories: abnormal components and attribute mis-binding. Abnormal components related artifacts mainly focus on three issues: distortion, omission and duplication regardless of the object type. For example, a popular omission is that human has fewer fingers or limbs without any specification. Attribute mis-binding, such as color, texture, and spatial position of different objects, also frequently emerges as a significant challenge for generative models. For example, the model may generate an object with the strange color that does not align with the common sense while the color is not mentioned in the text prompt. Finally, we also pay attention to lighting and blur, which is also a noteworthy problem causing the unreality of synthetic images.

\begin{table}[t]
\centering
\caption{\textbf{Sources of Prompts and Synthetic Images.} Prompts are sampled from image-caption datasets and user input. We employ diffusion model to generated images. }
\label{Distribution of our dataset}
\vspace{10pt}
\resizebox{1.0\columnwidth}{!}{
\begin{tabular}{ccc}
\toprule
    Prompt Source              & Generative Model                       & \#Images \\
\midrule
    ImageNet                   & Stable Diffusion v2.1                  & 100 \\
    MSCOCO                     & Stable Diffusion v2.1                  & 200 \\
    DrawBench                  & Stable diffusion v1.0                  & 40 \\
    DrawBench                  & Stable diffusion v1.4                  & 40 \\
    DrawBench                  & Stable diffusion v1.5                  & 40 \\
    DrawBench                  & Stable diffusion v2.0                  & 40 \\
    DrawBench                  & Stable diffusion v2.1                  & 40 \\
    ImageReward                & Stable Diffusion                       & 400 \\
    Midjourney Users           & Stable Diffusion v2.1                  & 210 \\
    DALLE-3 Users              & DALLE-3                                & 200 \\
\bottomrule
\end{tabular}
}
\end{table}

\subsubsection{Image Generation and Annotation}
\textbf{Prompt Collection and Image Generation.} To ensure the diversity of prompts and corresponding synthetic images, we collect and select prompts from three sources. Firstly, partial prompts are sampled from image-caption datasets, including ImageNet \cite{deng2009imagenet}, COCO captions \cite{chen2015microsoft}. Secondly, a number of prompts are user input of Midjourney and DALLE-3 \cite{betker2023improving}. Then, we use Stable Diffusion v2.1 \cite{rombach2021highresolution} to generate images. We also sample synthetic images collected by previous works, such as HPSv2 \cite{wu2023human} (We only utilize image generated from DrawBench \cite{saharia2022photorealistic}) and ImageReward \cite{xu2023imagereward}. Finally, we construct a dataset containing 1.3K text-image pairs, which is summarized in Tab.\ref{Distribution of our dataset}. 

\begin{figure}[t]
\centering
    \includegraphics[width=1.0\columnwidth]{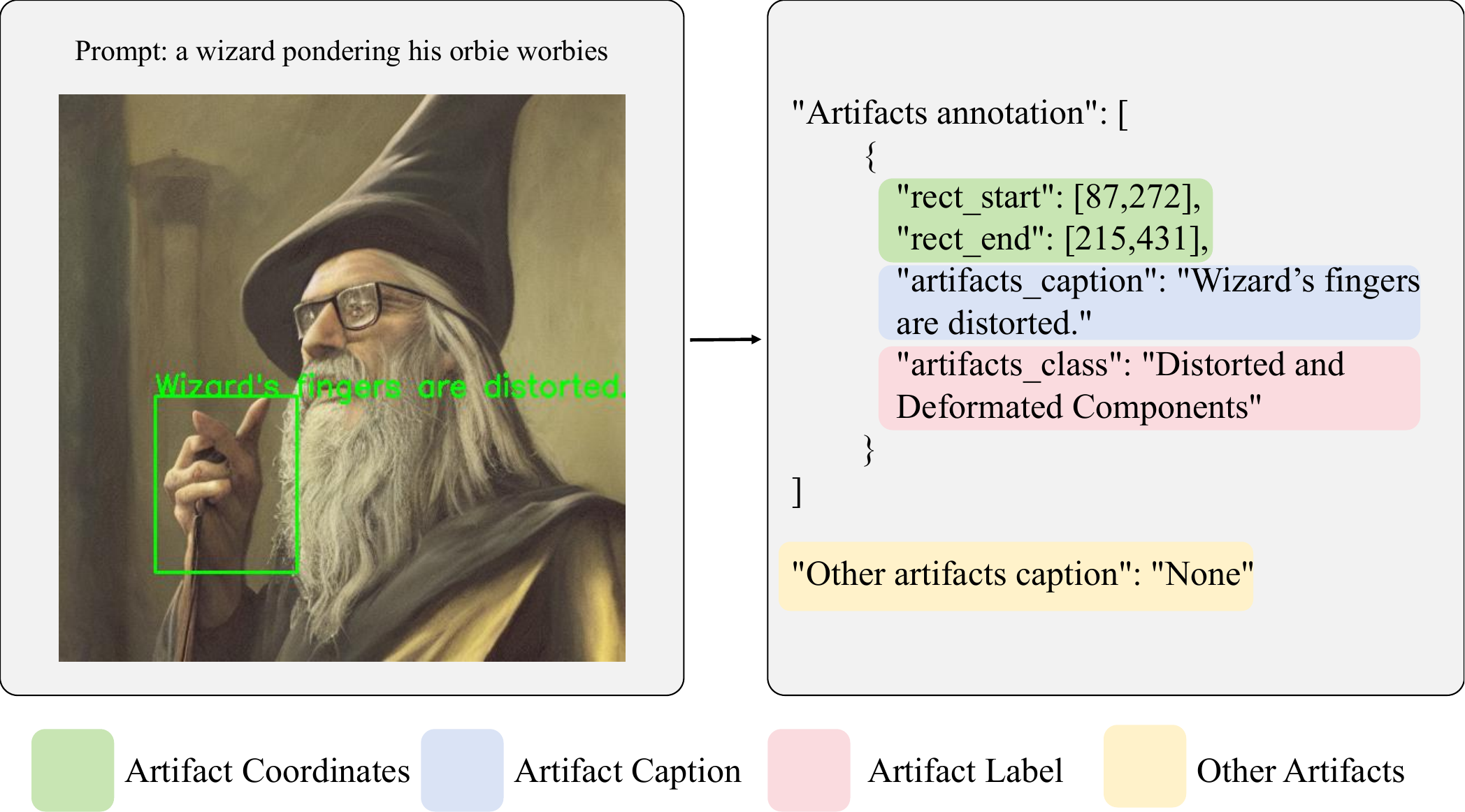}
\caption{\textbf{Our Annotation Paradigm.} Each synthetic image is annotated with coordinates, label and caption of artifacts.}
\label{fig:Annotation Surface}
\end{figure}

\begin{figure}[t]
\centering
    \includegraphics[width=1.0\columnwidth]{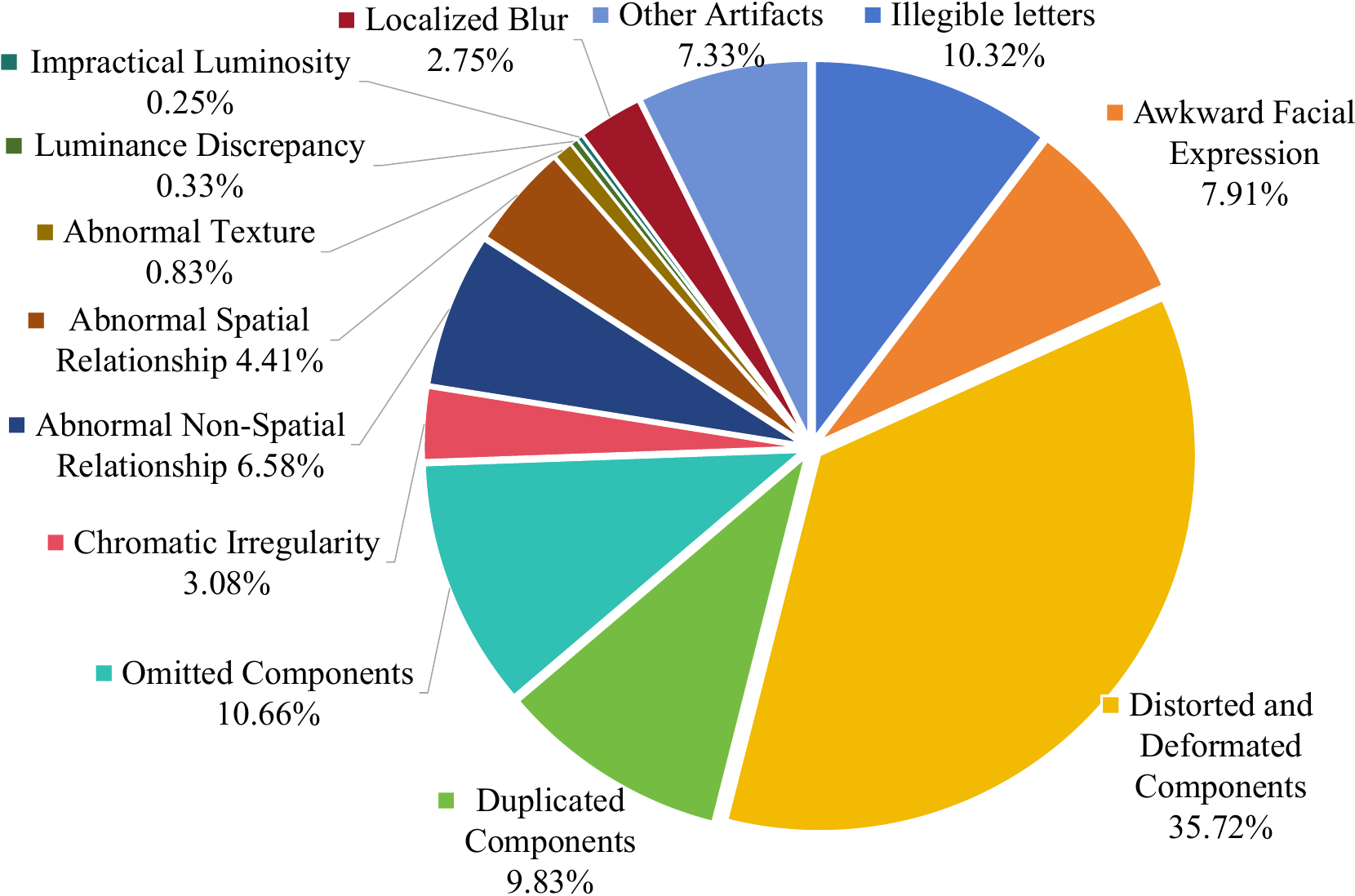}
\caption{\textbf{Distribution of Artifact Annotation.} Distortion, omission, and duplication are dominant artifacts while artifacts related to lighting are infrequent.}
\label{fig:artifacts_distribution}
\end{figure}

\begin{figure}[t]
\centering
    \includegraphics[width=1.0\columnwidth]{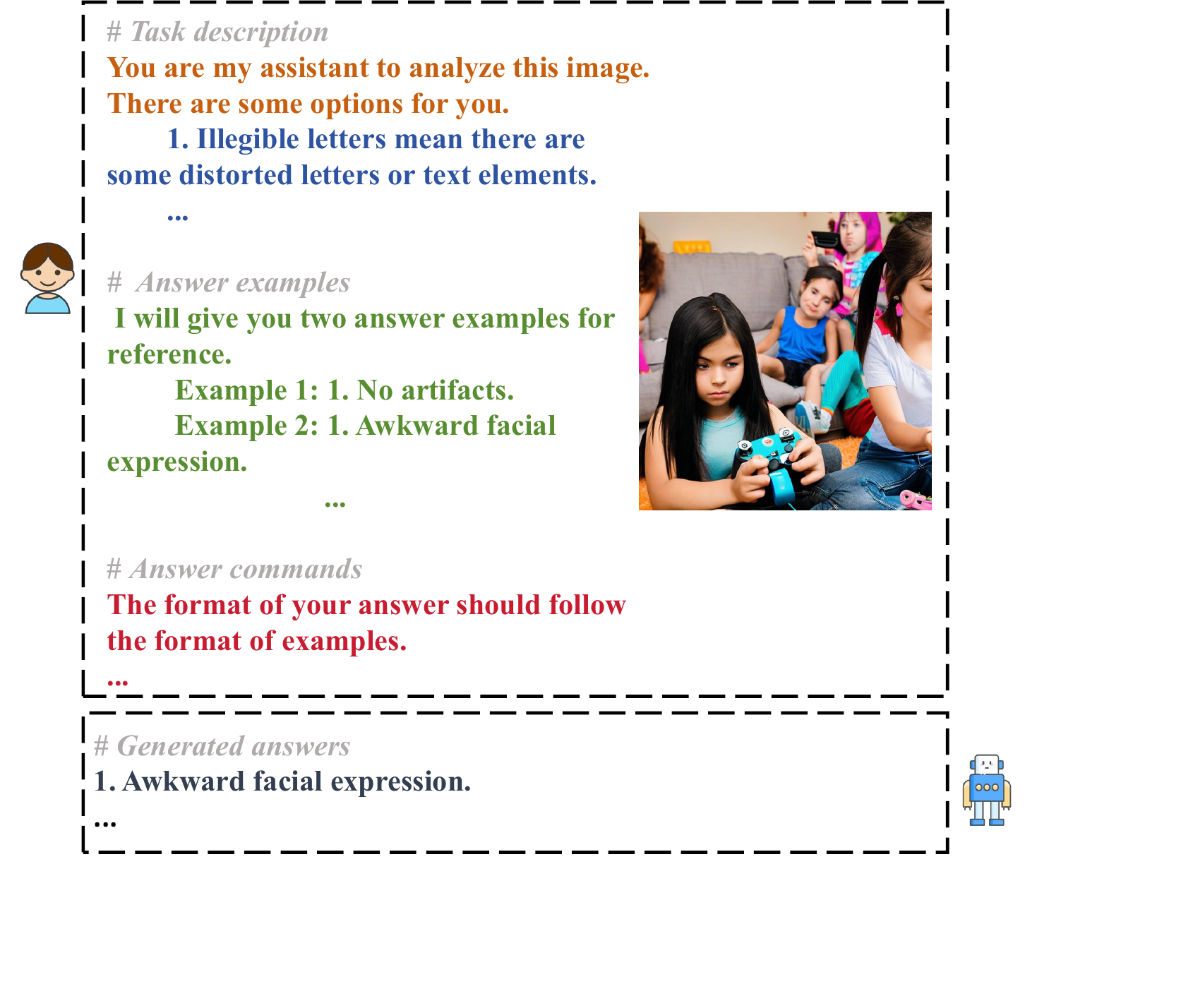}
\caption{\textbf{Question-Answer Template for Artifact Classification.} The template includes a set of options. Each option is accompanied with a detailed explanation. Reference answers are used to ensure the standardized format for response.}
\label{fig:Singal Convsersation}
\end{figure}

\textbf{Human Annotation and Analysis.} Previous annotation paradigms primarily focus on global scoring and ranking images \cite{xu2023imagereward,wu2023human,kirstain2023pick,10262331}, providing limited information for alleviating artifacts. Our annotation paradigm is shown in \cref{fig:Annotation Surface}. 
The annotation includes labels (see \cref{sec:artifacts taxonomy}), coordinates, and captions (detailed descriptions) of artifacts. Recognizing the challenge of locating all artifacts, we additionally utilize captions to describe other unspecified artifacts. Following the above annotation paradigm, we construct a dataset consisting of 768 images with artifacts and 542 images without artifacts, called SynArtifact-1K. The distribution of artifact annotation is illustrated in \cref{fig:artifacts_distribution}, showing that distortion is the predominant type of artifacts and luminosity is relatively infrequent. This imbalance could arise from both the weakness of generative model and the biased distribution of human prompts.

\subsection{Automatic Artifact Classification via Fine-tuning Vision-Language Model}
\label{sec:Automatic Artifact Detection with Vision-Language Model}
To fine-tune VLM as artifact classifier utilizing SynArtifact-1K,
we design a prompt template to transform various types of artifacts into instructions facilitating fine-tuning VLM, as shown in \cref{fig:Singal Convsersation}. In accordance with \cref{sec:artifacts taxonomy}, we consider 13 types of artifacts. Initially, we approach artifact classification as multiple binary classification tasks, designing a question for each kind of artifact. Every answer can only be \textit{``Yes"} or \textit{``No"}. However, since a synthetic image typically contains a few kinds of artifacts, most answers of questions are \textit{``No"}, leading to imbalance within answers. After fine-tuning, we observe that VLM tends to answer \textit{``No"}. To address this imbalance, we modify our approach to adopt a single question-answer pair for artifact classification, only answering types of artifacts in synthetic image rather than posing questions about each kind of artifact individually. Based above analysis, we consider artifact classification as a multi-label classification task. Our text prompt is designed as follow:
\begin{align*}
&\textit{Task description} \xrightarrow{} \textit{Presenting all options}\\
&\xrightarrow{} \textit{Answer examples} \xrightarrow{} \textit{Answer commands}
\end{align*}
By presenting all available options with detailed explanations, VLM selects one or more options for response. Additionally, we also provide two reference answers and answer commands to ensure the standardized response format and avoid in-context conflicts, as shown in \cref{fig:Singal Convsersation}.

In our work, we employ LLaVA \cite{liu2023improvedllava, liu2023llava} $\Phi_{\text{VLM}}(\cdot, \cdot)$ utilized for artifact classification, which contains a vision encoder $\Phi_{\theta_{v}}(\cdot)$, a pre-trained large language model $\Phi_{\theta_{l}}(\cdot)$ and a vision-language projector $\Phi_{\theta_{c}}(\cdot)$. Given a synthetic image $x_{v}$ and a question $x_{q}$, the vision encoder extracts image features $f_{v}=\Phi_{\theta_{v}}(x_{v})$ and the vision-language projector convert image features into the word embedding space. Subsequently, the vision and language tokens are concatenated and fed into LLM to generate answers:
\begin{equation}
    \tilde{x}_{a} = \Phi_{\theta_{l}}(x_{q}, \Phi_{\theta_{c}}(f_{v}))
\end{equation}
The cross entropy loss is calculated between generated answers $\tilde{x}_{a}$ and ground truth $x_{a}$ to update parameters of vision-language projector $\theta_{c}$ and LLM $\theta_{l}$:
\begin{equation}
    \theta^{*}_{l}, \theta^{*}_{c} = \operatorname*{argmin}_{\theta_{l}, \theta_{c}} \mathcal{L}_{CE}(\tilde{x}_{a}, x_{a})
\end{equation}

\begin{table*}[ht]
\centering
\caption{\textbf{Artifact Classification Experiments.} We ablate multiple design choices, including fine-tuning dataset, weight initialization and updating LLM. In all experiments, the vision encoder is frozen while the vision-language projector is tuned. \textbf{Stage 1} denotes we load the \textbf{pre-trained} weights of LLaVA-v1.5 as the initialization. \textbf{Stage 2} denotes we load the \textbf{visual instruction-tuned} weights of LLaVA-v1.5 as the initialization. Note that fine-tuning on LLaVA-v1.5-mix665K results in the vanilla LLaVA-v1.5, which is the baseline.}
\label{tab:ablation}
\vspace{5pt}
\resizebox{1.0\linewidth}{!}{
\begin{tabular}{c|c|c|c|cccc}
\toprule
\textbf{Fine-tuning Dataset}  &\textbf{Weight Initialization} & \textbf{Updating LLM}     & \textbf{Categories}      & \textbf{Accuracy}    & \textbf{Precision}   & \textbf{Recall}    & \textbf{F1 Score}  \\ 
\midrule
\multirow{3}{*}{LLaVA-v1.5-mix665K}   & \multirow{3}{*}{Stage 1} & \multirow{3}{*}{\checkmark}                             
                                                                        & All                      & \cellcolor{gray!25}20.00       & \cellcolor{gray!25}25.28       & \cellcolor{gray!25}22.14     & \cellcolor{gray!25}23.01   \\
&        &                                                              & Distortion               & -       & -      & -      & -    \\ 

&        &                                                              & No artifacts              & \cellcolor{gray!12}55.09       & \cellcolor{gray!12}46.67       & \cellcolor{gray!12}43.75     & \cellcolor{gray!12}45.16   \\ 
\midrule

\multirow{9}{*}{SynArtifact-1K}       


& \multirow{3}{*}{Stage 1}       & \multirow{3}{*}{\ding{55}}                                   
                                                                        & All                      & \cellcolor{gray!25}\textbf{45.66}     & \cellcolor{gray!25}\textbf{54.40}         & \cellcolor{gray!25}\textbf{51.84}     &\cellcolor{gray!25} \textbf{52.02}   \\

&        &                                                              & Distortion               & \textbf{71.70}     & \textbf{63.08}        & \textbf{44.57}      & \textbf{52.23}   \\ 

&        &                                                              & No artifacts              & \cellcolor{gray!12}\textbf{65.66}     & \cellcolor{gray!12}\textbf{55.74}        & \cellcolor{gray!12}\textbf{91.07}      & \cellcolor{gray!12}\textbf{69.15}   \\ 
\cmidrule(lr){2-8}

& \multirow{3}{*}{Stage 1}       & \multirow{3}{*}{\checkmark}                                             
                                                                         & All                     & \cellcolor{gray!25}36.23     & \cellcolor{gray!25}48.87        & \cellcolor{gray!25}47.30      & \cellcolor{gray!25}46.50   \\
&        &                                                               & Distortion              & 62.26     & 46.92        & 66.30      & 54.95   \\ 

&        &                                                               & No artifacts             & \cellcolor{gray!12}63.40     & \cellcolor{gray!12}55.56        & \cellcolor{gray!12}66.96     & \cellcolor{gray!12}60.73   \\ 

\cmidrule(lr){2-8}

& \multirow{3}{*}{Stage 2}       & \multirow{3}{*}{\ding{55}}                                             
                                                                         & All               & \cellcolor{gray!25}34.34       & \cellcolor{gray!25}39.87       & \cellcolor{gray!25}42.30     & \cellcolor{gray!25}40.10\\
 &                                                   &        & Distortion        & 57.74       & 36.84      & 30.43     & 33.33\\
 &                                                 &          & No artifacts       & \cellcolor{gray!12}53.58       & \cellcolor{gray!12}46.99       & \cellcolor{gray!12}76.79     & \cellcolor{gray!12}58.31  \\
        
\bottomrule

\end{tabular}
}
\end{table*}

\subsection{Artifact Classification as AI Feedback for Alleviating Artifact}
\label{sec:Detection4RL}
Based on the developed artifact alleviation pipeline (see \cref{fig:pipeline}), we further investigate 
reinforcement learning from AI feedback to improve the quality of synthetic images. For this purpose, we design a reward function reflecting the defined ``human'' perception of artifacts in images. This reward function is then employed to fine-tune the text-to-image generative model (i.e., latent diffusion model \cite{rombach2021highresolution}) through online reinforcement learning to reduce the occurrence of artifacts in synthetic images. Several works  \cite{black2023training,fan2023dpok, wallace2023diffusion} have demonstrated the effectiveness of policy gradient optimization-based reinforcement learning for diffusion model fine-tuning, which provides better maximization of human reward  \cite{fan2023dpok} and preference alignment  \cite{wallace2023diffusion}.

Specifically in our application, we adopt the Denoising Diffusion Policy Optimization (DDPO) \cite{black2023training} to fine-tune Stable Diffusion  \cite{rombach2022high}. This involves a delicate reward function design of converting the discrete artifact classifier output into a continuous and informative artifact-aware reward for an effective estimate of policy gradient for optimization. 

\begin{equation}
    \begin{aligned}
        \nabla_\theta \mathcal{J} &= \mathbb{E}[\sum_{t=0}^T \frac{p_\theta\left({x}_{t-1} \mid {x}_t, {c}\right)}{p_{\theta_{\text{old }}}\left({x}_{t-1} \mid {x}_t, {c}\right)}\\
        &\nabla_\theta \log p_\theta\left({x}_{t-1} \mid {x}_t, {c}\right) r\left({x}_v, {c}\right)]
    \label{eq:ddpo_1}
    \end{aligned}
\end{equation}

As indicated in \cref{eq:ddpo_1}, the policy gradient $\nabla_\theta \mathcal{J}$ is estimated with an expectation over multiple denoising steps ${x}_{t}$ in diffusion process with importance sampling. The reward function $r\left({x}_v, {c}\right)$ is designed as in \cref{eq:reward}:

\begin{equation}
    \begin{aligned}
    r({x}_v, {c}) &= \text{BertScore}(''\text{No artifacts}'', \Phi_{\text{VLM}}({x}_v,{x}_q)) \\
    & - \alpha \sum_{k}^{N} \text{BertScore}(s_k, \Phi_{\text{VLM}}({x}_v,{x}_q)) + \beta,
    \label{eq:reward}
\end{aligned}
\end{equation}

where \( \Phi_{\text{VLM}} \) is our artifact classifier, and \( {x}_v \) is the denoised synthetic image and \( {x}_q \) is question for our artifact classifier during fine-tuning. We adopt \emph{BertScore} \cite{zhang2019bertscore} to measure the similarity between the output of our artifact classifier $\Phi_{\text{VLM}}({x}_v,{x}_q)$ and a set of pre-defined phrases of artifact class $s_k$ (see Section \ref{sec:artifacts taxonomy}), where $k$ is the index of artifact category. The reward function encourages the generative model to produce images with ``\textit{No artifacts}'' while penalizing the overall weighted sum of each artifact. The weight coefficient \( \alpha \) and offset \( \beta \) are set to ensure a positive reward function.

\section{Experiments}
\label{sec:experiment}
\subsection{Dataset \& Setting} We divide SynArtifact-1K into a training set, comprising 1045 annotated images, and a testing set, containing 265 annotated images. 
We employ LLaVA-v1.5-7B \cite{liu2023improvedllava, liu2023llava} and fine-tune it for experiments. For artifact classification, we conduct comprehensive ablation studies on fine-tuning dataset, weight initialization, updating LLM. For artifact classification, the batch size is set to be 16.
For fine-tuning Stable Diffusion v1.5, we adopt the LoRA \cite{hu2021lora} strategy and the prompts in ImageNet-animal\footnote{Detailed prompt list is available at \url{https://github.com/kvablack/ddpo-pytorch}}. We set the learning rate 3e-4 and the batch size 24.  

\subsection{Artifact Classification Results} 
\paragraph{Comparison to Baseline.} Recognizing the potential presence of multiple types of artifacts in a synthetic image, we consider the artifact classification as a multi-label classification task (see \cref{sec:Automatic Artifact Detection with Vision-Language Model}). The label space includes 13 kinds of artifacts (see \cref{fig:Artifact Taxonomy}) and \textit{``No artifacts"}. Then, we evaluate the ability to identify artifacts on the testing set. As shown in \cref{tab:ablation}, LLaVA, fine-tuned on SynArtifact-1K, demonstrates a remarkable ability in artifact classification and the accuracy reaches up to 45.66\%, achieving a notable improvement of 25.66\%, compared with the baseline LLaVA-v1.5 which is fine-tuned on LLaVA-v1.5-mix665K. Lacking synthetic images in LLaVA-v1.5-mix665K, LLaVA tends to categorize an input image as a normal image without artifacts. Instead, through fine-tuning on SynArtifact-1K, the accuracy of \textit{``No artifacts"} further improves by 10.57\% and showcases the enhanced ability to identify artifacts.
Considering distortion is the most prevalent artifact, we also present the performance on this category in \cref{tab:ablation}. The baseline LLaVA-v1.5 model performs extremely bad in identifying distortion in synthetic images, while ours achieve 52.23\% F1 Score.

\paragraph{Updating LLM Parameters.} As illustrated in \cref{tab:ablation}, updating LLM parameters causes much worse  performance. The potential reason is that the presented SynArtifact-1K dataset is small, tuning LLM on the small set will cause overfitting and thus worse performance. 
\begin{figure*}[t]
\centering
    \includegraphics[width=1.0\textwidth]{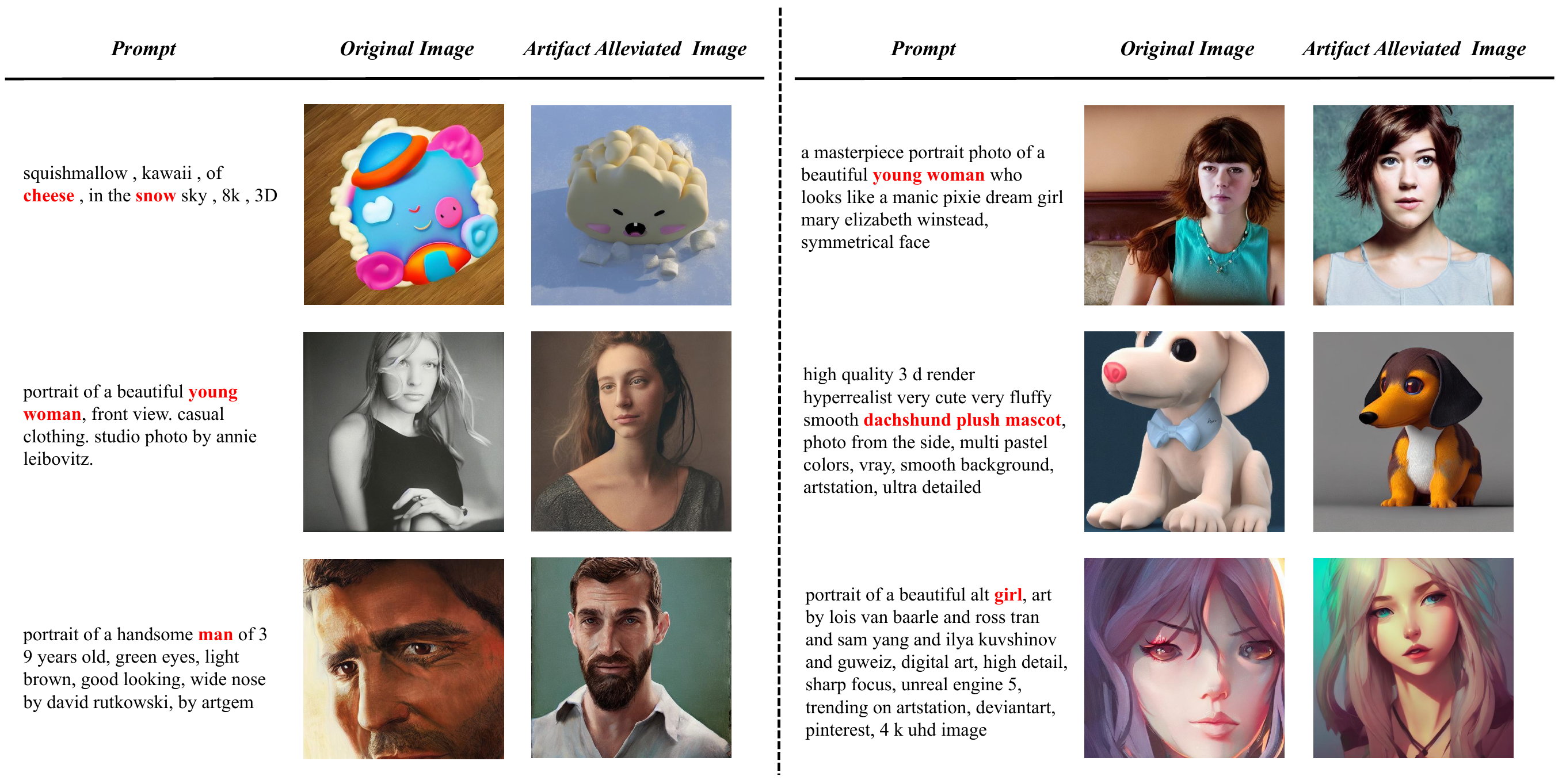}
\caption{\textbf{Visualization of Samples in Reinforcement Learning from Artifact Classification.} Each example contains the prompt, synthetic image and artifact-alleviated image, showing that artifacts in synthetic image have been alleviated through DDPO.}
\label{fig:ddpo_vis}
\end{figure*}

\paragraph{Weight Initialization.} Considering that LLaVA training process consists of two stages: feature alignment, i.e. pre-training (\textbf{Stage 1}) and visual instruction tuning (\textbf{Stage 2}), we conduct a comprehensive investigation on the impact of weight initialization for artifact classification. As shown in \cref{tab:ablation}, we observe that fine-tuning LLaVA utilizing SynArtifact-1K based on Stage 1 achieves better performance. It is worth noting that fine-tuned model based on Stage 2 achieves a recall of 30.43\% for \textit{``Distortion''} but for a recall of 76.79\% for \textit{``No artifacts"}, indicating that model still tend to misclassify a synthetic image with distortion into a synthetic image without distorted components. 

\subsection{Primary Experiment on Artifact Detection}
\begin{figure}[t]
\centering
    \includegraphics[width=0.96\columnwidth]{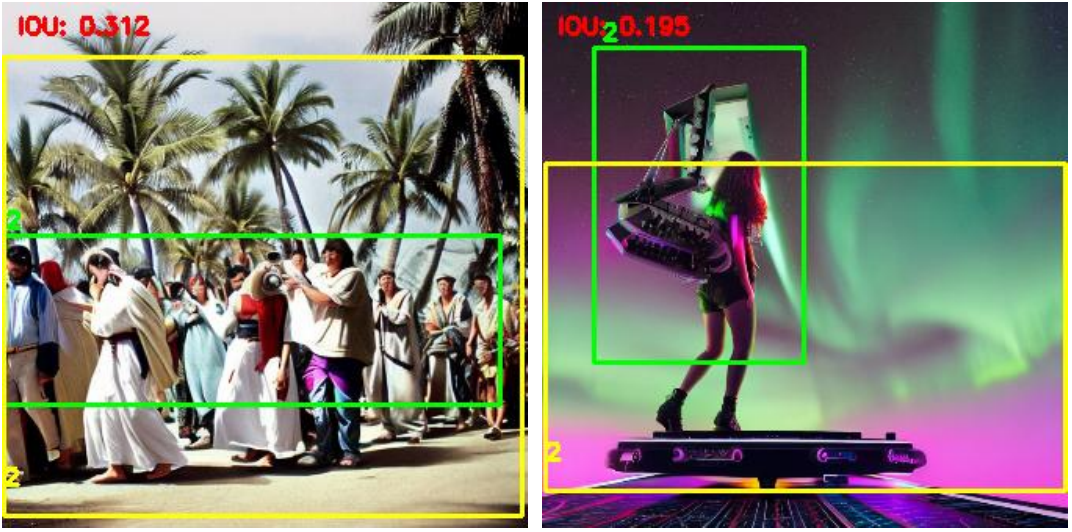}
\caption{\textbf{Artifact Detection Results.} We present artifact detection results, with the ground truth, i.e. the human annotations in \textbf{Green}, and the detection results in \textbf{Yellow}. IOU is highlighted on the top-left of the image. The number at the left lower corner in the bounding box is the artifact category index, and category \textit{``2"} denotes \textit{``Distorted and deformated components''.} }
\label{fig:detection_results}
\end{figure}

To locate artifacts, we further attempt to perform artifact detection. The chat template for artifact detection and implement details can be seen in \cref{sec:appendix_Instruction_Artifact_Detection}. The primary results of artifact detection are shown in \cref{fig:detection_results}. LLaVA shows the primary ability to locate artifacts in synthetic images through fine-tuning on Synthetic-1K. To evaluate detection results quantitatively, we calculate IOU (Intersection Over Union) between the predicted bounding box and ground truth containing human annotated bounding boxes of the same kinds of artifacts. However, we observe that LLaVA struggles with precise artifact localization. The fined-tuned LLaVA tends to predict a large bounding box to include multiple artifacts. The potential reasons are that LLaVA lacks the ability of localization inherently and the presented SynArtifact-1K dataset is small. This observation underscores the challenge and potential of artifact detection task, which can develop more explainable quality evaluation metrics for synthetic images and provide valuable insights for optimizing diffusion models. In the future, Vision-Language Models with powerful visual grounding ability, such as Shikra \cite{chen2023shikra} and MiniGPT-v2 \cite{chen2023minigptv2}, are also attempting. 

\begin{figure}[t]
\centering
    \includegraphics[width=1.0\columnwidth]{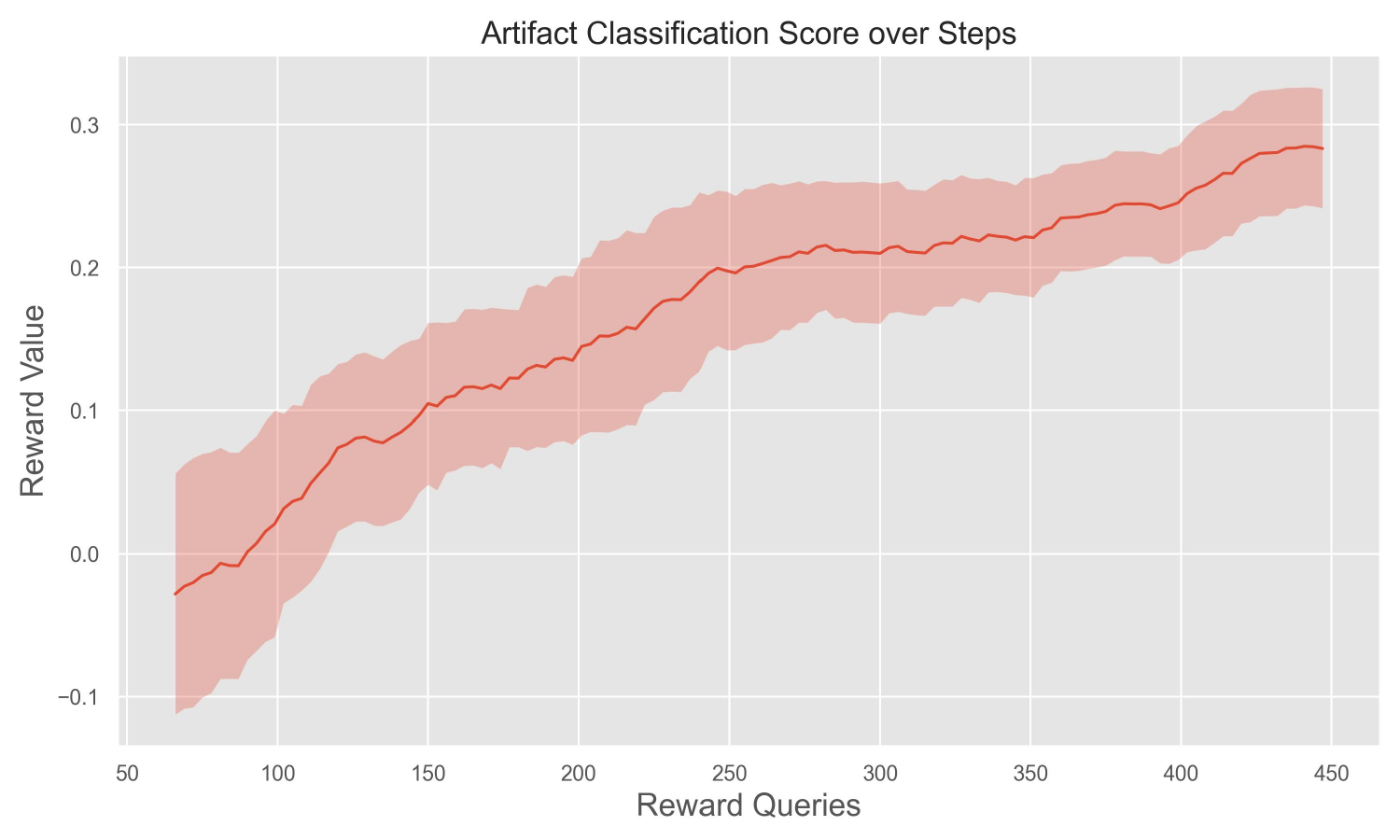}
\caption{\textbf{Improving Artifact Classification Reward.} Synthetic images evaluated with the artifact classifier obtain increasing rewards with more queries from the artifact classifier, indicating fewer artifacts are presented.}
\label{fig:ddpo_reward}
\end{figure}

\subsection{Improved Generative Model with Artifact Alleviation}
We further demonstrate the effectiveness and applicability of the artifact classifier for fine-tuning diffusion model. With the designed reward feedback (\cref{sec:Detection4RL}), Stable Diffusion produces higher quality synthetic images with higher artifact reward as shown in \cref{fig:ddpo_reward}. Note that our artifact classification reward increases quickly, indicating the efficiency of the DDPO strategy and the informativeness of the designed artifact score in reflecting the artifact presented in synthetic images.

\paragraph{Visualization.} To better illustrate the effectiveness of reinforcement learning from the artifact classifier, comparison between synthetic images produced by base and fine-tuned diffusion model are present in \cref{fig:ddpo_vis}. We can observe clear alleviation and removal of several notable artifacts (i.e., distorted limbs, duplicated components, out of frame, and awkward facial expression) during the fine-tuning process. For example, the man who is out of frame (row 3, column 1) has been moved to the center of image and the dog whose head is incomplete (row 2, column 2) has been refined to a dog with perfect details.  

\begin{figure}[t]
\centering
    \includegraphics[width=1.0\columnwidth]{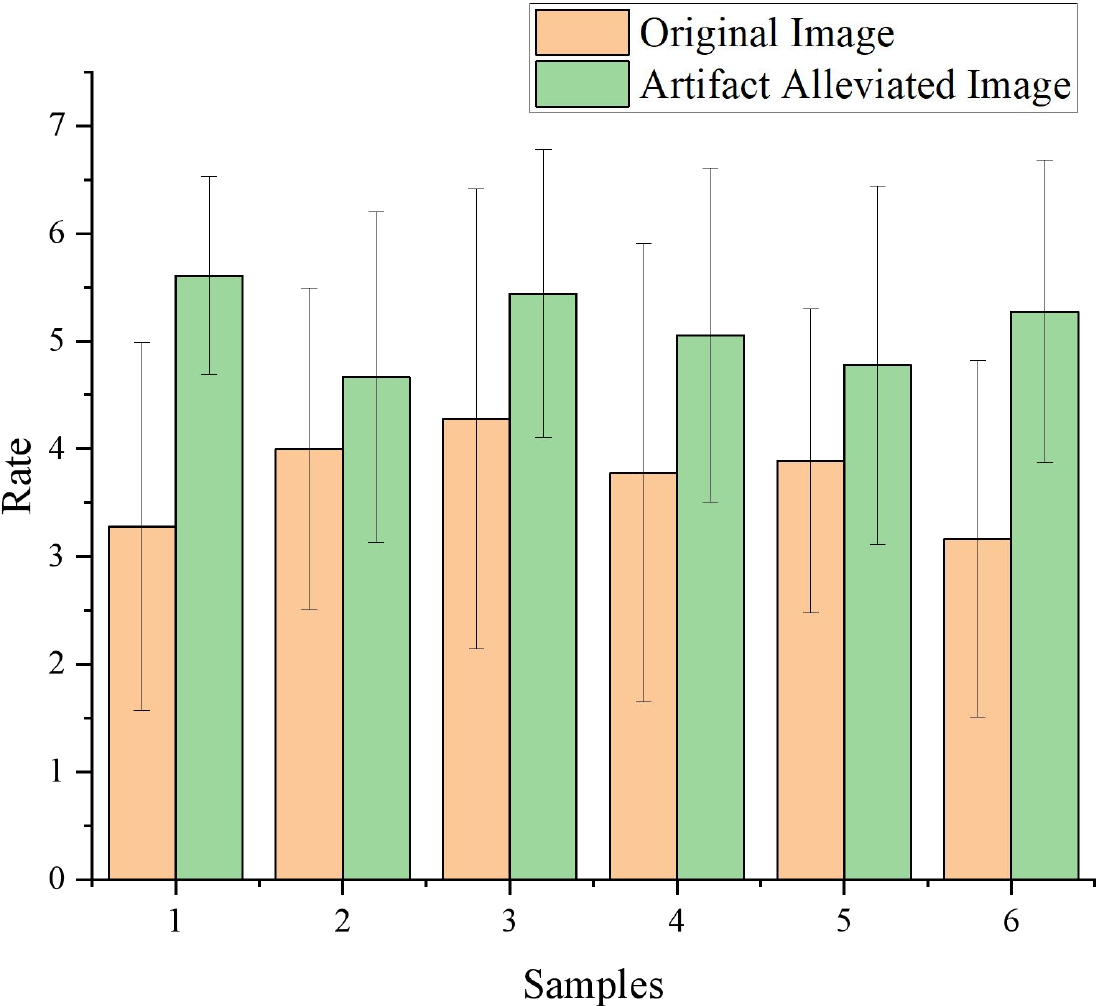}
\caption{\textbf{User Study.} We present mean (\textbf{Histogram}) and variance (\textbf{Boxplot}) of human rate for original image and artifact alleviated image. Artifact alleviated images beat original images.}
\label{fig:user_study}
\end{figure}

\paragraph{User Study.} To quantitatively evaluate artifact alleviation effects leveraging RLAIF, we recruit 18 well-educated volunteers to rate the synthetic images from 1 to 7 for the image quality. The results are shown in \cref{fig:user_study}. We can find that synthetic images generated by our tuned diffusion model can better align with human preference, indicating that RLAIF remains a beneficial impact in alleviating artifacts in synthetic images.

\section{Limitations}
\label{sec:limitations}
In this section, we discuss some limitations of our work. SynArtifact-1K contains 1.3k text-image pairs and we recognize better artifact classification and alleviation performance could be achieved with the larger scale of dataset. 
While LLaVA serves as our foundation model that demonstrates artifact classification ability through fine-tuning, the artifact detection ability still needs improvements because LLaVA inherently lacks the detection ability. In the future work, the model with powerful detection ability is worth attempting.

\section{Conclusion}
\label{sec:conclusion}
In this paper, we focus on artifacts in synthetic images and address two key issues: artifact classification and alleviation. We construct a systematic artifact taxonomy involving 13 kinds of common artifacts, and create the first image-with-artifact dataset annotated with artifact categories, descriptions and coordinates, named SynArtifact-1K. A Vision Language Model is fine-tuned on SynArtifact-1K to automatically classify artifacts, and further used to refine the generative model in the manner of RLAIF. Extensive experiments of visualization and user study reveal that our method can effectively improve the quality of synthetic images. 

\bibliography{example_paper}

\begin{thebibliography}{44}
\providecommand{\natexlab}[1]{#1}
\providecommand{\url}[1]{\texttt{#1}}
\expandafter\ifx\csname urlstyle\endcsname\relax
  \providecommand{\doi}[1]{doi: #1}\else
  \providecommand{\doi}{doi: \begingroup \urlstyle{rm}\Url}\fi

\bibitem[Bai et~al.(2022)Bai, Kadavath, Kundu, Askell, Kernion, Jones, Chen, Goldie, Mirhoseini, McKinnon, et~al.]{bai2022constitutional}
Bai, Y., Kadavath, S., Kundu, S., Askell, A., Kernion, J., Jones, A., Chen, A., Goldie, A., Mirhoseini, A., McKinnon, C., et~al.
\newblock Constitutional ai: Harmlessness from ai feedback.
\newblock \emph{arXiv preprint arXiv:2212.08073}, 2022.

\bibitem[Betker et~al.(2023)Betker, Goh, Jing, Brooks, Wang, Li, Ouyang, Zhuang, Lee, Guo, et~al.]{betker2023improving}
Betker, J., Goh, G., Jing, L., Brooks, T., Wang, J., Li, L., Ouyang, L., Zhuang, J., Lee, J., Guo, Y., et~al.
\newblock Improving image generation with better captions.
\newblock \emph{Computer Science. https://cdn. openai. com/papers/dall-e-3. pdf}, 2:\penalty0 3, 2023.

\bibitem[Black et~al.(2023)Black, Janner, Du, Kostrikov, and Levine]{black2023training}
Black, K., Janner, M., Du, Y., Kostrikov, I., and Levine, S.
\newblock Training diffusion models with reinforcement learning.
\newblock \emph{arXiv preprint arXiv:2305.13301}, 2023.

\bibitem[Chen et~al.(2023{\natexlab{a}})Chen, Zhu, Shen, Li, Liu, Zhang, Krishnamoorthi, Chandra, Xiong, and Elhoseiny]{chen2023minigptv2}
Chen, J., Zhu, D., Shen, X., Li, X., Liu, Z., Zhang, P., Krishnamoorthi, R., Chandra, V., Xiong, Y., and Elhoseiny, M.
\newblock Minigpt-v2: large language model as a unified interface for vision-language multi-task learning.
\newblock \emph{arXiv preprint arXiv:2310.09478}, 2023{\natexlab{a}}.

\bibitem[Chen et~al.(2023{\natexlab{b}})Chen, Zhang, Zeng, Zhang, Zhu, and Zhao]{chen2023shikra}
Chen, K., Zhang, Z., Zeng, W., Zhang, R., Zhu, F., and Zhao, R.
\newblock Shikra: Unleashing multimodal llm's referential dialogue magic.
\newblock \emph{arXiv preprint arXiv:2306.15195}, 2023{\natexlab{b}}.

\bibitem[Chen et~al.(2015)Chen, Fang, Lin, Vedantam, Gupta, Doll{\'a}r, and Zitnick]{chen2015microsoft}
Chen, X., Fang, H., Lin, T.-Y., Vedantam, R., Gupta, S., Doll{\'a}r, P., and Zitnick, C.~L.
\newblock Microsoft coco captions: Data collection and evaluation server.
\newblock \emph{arXiv preprint arXiv:1504.00325}, 2015.

\bibitem[Chen(2023)]{chen2023x}
Chen, Y.
\newblock X-iqe: explainable image quality evaluation for text-to-image generation with visual large language models.
\newblock \emph{arXiv preprint arXiv:2305.10843}, 2023.

\bibitem[Cho et~al.(2023)Cho, Zala, and Bansal]{Cho2023VPT2I}
Cho, J., Zala, A., and Bansal, M.
\newblock Visual programming for text-to-image generation and evaluation.
\newblock In \emph{NeurIPS}, 2023.

\bibitem[Deng et~al.(2009)Deng, Dong, Socher, Li, Li, and Fei-Fei]{deng2009imagenet}
Deng, J., Dong, W., Socher, R., Li, L.-J., Li, K., and Fei-Fei, L.
\newblock Imagenet: A large-scale hierarchical image database.
\newblock In \emph{2009 IEEE conference on computer vision and pattern recognition}, pp.\  248--255. Ieee, 2009.

\bibitem[Fan et~al.(2023)Fan, Watkins, Du, Liu, Ryu, Boutilier, Abbeel, Ghavamzadeh, Lee, and Lee]{fan2023dpok}
Fan, Y., Watkins, O., Du, Y., Liu, H., Ryu, M., Boutilier, C., Abbeel, P., Ghavamzadeh, M., Lee, K., and Lee, K.
\newblock Dpok: Reinforcement learning for fine-tuning text-to-image diffusion models.
\newblock \emph{arXiv preprint arXiv:2305.16381}, 2023.

\bibitem[Ghosh et~al.(2023)Ghosh, Hajishirzi, and Schmidt]{ghosh2023geneval}
Ghosh, D., Hajishirzi, H., and Schmidt, L.
\newblock Geneval: An object-focused framework for evaluating text-to-image alignment.
\newblock \emph{arXiv preprint arXiv:2310.11513}, 2023.

\bibitem[Hessel et~al.(2021)Hessel, Holtzman, Forbes, Bras, and Choi]{hessel2021clipscore}
Hessel, J., Holtzman, A., Forbes, M., Bras, R.~L., and Choi, Y.
\newblock Clipscore: A reference-free evaluation metric for image captioning.
\newblock \emph{arXiv preprint arXiv:2104.08718}, 2021.

\bibitem[Heusel et~al.(2017)Heusel, Ramsauer, Unterthiner, Nessler, and Hochreiter]{heusel2017gans}
Heusel, M., Ramsauer, H., Unterthiner, T., Nessler, B., and Hochreiter, S.
\newblock Gans trained by a two time-scale update rule converge to a local nash equilibrium.
\newblock \emph{Advances in neural information processing systems}, 30, 2017.

\bibitem[Ho et~al.(2020)Ho, Jain, and Abbeel]{ho2020denoising}
Ho, J., Jain, A., and Abbeel, P.
\newblock Denoising diffusion probabilistic models.
\newblock \emph{Advances in neural information processing systems}, 33:\penalty0 6840--6851, 2020.

\bibitem[Hu et~al.(2021)Hu, Shen, Wallis, Allen-Zhu, Li, Wang, Wang, and Chen]{hu2021lora}
Hu, E.~J., Shen, Y., Wallis, P., Allen-Zhu, Z., Li, Y., Wang, S., Wang, L., and Chen, W.
\newblock Lora: Low-rank adaptation of large language models.
\newblock \emph{arXiv preprint arXiv:2106.09685}, 2021.

\bibitem[Hu et~al.(2023)Hu, Liu, Kasai, Wang, Ostendorf, Krishna, and Smith]{hu2023tifa}
Hu, Y., Liu, B., Kasai, J., Wang, Y., Ostendorf, M., Krishna, R., and Smith, N.~A.
\newblock Tifa: Accurate and interpretable text-to-image faithfulness evaluation with question answering.
\newblock \emph{arXiv preprint arXiv:2303.11897}, 2023.

\bibitem[Huang et~al.(2023)Huang, Sun, Xie, Li, and Liu]{huang2023t2i}
Huang, K., Sun, K., Xie, E., Li, Z., and Liu, X.
\newblock T2i-compbench: A comprehensive benchmark for open-world compositional text-to-image generation.
\newblock \emph{arXiv preprint arXiv:2307.06350}, 2023.

\bibitem[Kirstain et~al.(2023)Kirstain, Polyak, Singer, Matiana, Penna, and Levy]{kirstain2023pick}
Kirstain, Y., Polyak, A., Singer, U., Matiana, S., Penna, J., and Levy, O.
\newblock Pick-a-pic: An open dataset of user preferences for text-to-image generation.
\newblock \emph{arXiv preprint arXiv:2305.01569}, 2023.

\bibitem[Lee et~al.(2023)Lee, Phatale, Mansoor, Lu, Mesnard, Bishop, Carbune, and Rastogi]{lee2023rlaif}
Lee, H., Phatale, S., Mansoor, H., Lu, K., Mesnard, T., Bishop, C., Carbune, V., and Rastogi, A.
\newblock Rlaif: Scaling reinforcement learning from human feedback with ai feedback.
\newblock \emph{arXiv preprint arXiv:2309.00267}, 2023.

\bibitem[Li et~al.(2023)Li, Zhang, Wu, Sun, Min, Liu, Zhai, and Lin]{10262331}
Li, C., Zhang, Z., Wu, H., Sun, W., Min, X., Liu, X., Zhai, G., and Lin, W.
\newblock Agiqa-3k: An open database for ai-generated image quality assessment.
\newblock \emph{IEEE Transactions on Circuits and Systems for Video Technology}, pp.\  1--1, 2023.
\newblock \doi{10.1109/TCSVT.2023.3319020}.

\bibitem[Li et~al.(2022)Li, Li, Xiong, and Hoi]{li2022blip}
Li, J., Li, D., Xiong, C., and Hoi, S.
\newblock Blip: Bootstrapping language-image pre-training for unified vision-language understanding and generation.
\newblock In \emph{International Conference on Machine Learning}, pp.\  12888--12900. PMLR, 2022.

\bibitem[Liu et~al.(2023{\natexlab{a}})Liu, Li, Li, and Lee]{liu2023improvedllava}
Liu, H., Li, C., Li, Y., and Lee, Y.~J.
\newblock Improved baselines with visual instruction tuning, 2023{\natexlab{a}}.

\bibitem[Liu et~al.(2023{\natexlab{b}})Liu, Li, Wu, and Lee]{liu2023llava}
Liu, H., Li, C., Wu, Q., and Lee, Y.~J.
\newblock Visual instruction tuning, 2023{\natexlab{b}}.

\bibitem[Mann et~al.(2020)Mann, Ryder, Subbiah, Kaplan, Dhariwal, Neelakantan, Shyam, Sastry, Askell, Agarwal, et~al.]{mann2020language}
Mann, B., Ryder, N., Subbiah, M., Kaplan, J., Dhariwal, P., Neelakantan, A., Shyam, P., Sastry, G., Askell, A., Agarwal, S., et~al.
\newblock Language models are few-shot learners.
\newblock \emph{arXiv preprint arXiv:2005.14165}, 2020.

\bibitem[Ouyang et~al.(2022)Ouyang, Wu, Jiang, Almeida, Wainwright, Mishkin, Zhang, Agarwal, Slama, Ray, et~al.]{ouyang2022training}
Ouyang, L., Wu, J., Jiang, X., Almeida, D., Wainwright, C.~L., Mishkin, P., Zhang, C., Agarwal, S., Slama, K., Ray, A., et~al.
\newblock Training language models to follow instructions with human feedback, 2022.
\newblock \emph{URL https://arxiv. org/abs/2203.02155}, 13, 2022.

\bibitem[Radford et~al.(2021)Radford, Kim, Hallacy, Ramesh, Goh, Agarwal, Sastry, Askell, Mishkin, Clark, et~al.]{radford2021learning}
Radford, A., Kim, J.~W., Hallacy, C., Ramesh, A., Goh, G., Agarwal, S., Sastry, G., Askell, A., Mishkin, P., Clark, J., et~al.
\newblock Learning transferable visual models from natural language supervision.
\newblock In \emph{International conference on machine learning}, pp.\  8748--8763. PMLR, 2021.

\bibitem[Ramesh et~al.(2022)Ramesh, Dhariwal, Nichol, Chu, and Chen]{ramesh2022hierarchical}
Ramesh, A., Dhariwal, P., Nichol, A., Chu, C., and Chen, M.
\newblock Hierarchical text-conditional image generation with clip latents, 2022.
\newblock \emph{URL https://arxiv. org/abs/2204.06125}, 7, 2022.

\bibitem[Rombach et~al.(2021)Rombach, Blattmann, Lorenz, Esser, and Ommer]{rombach2021highresolution}
Rombach, R., Blattmann, A., Lorenz, D., Esser, P., and Ommer, B.
\newblock High-resolution image synthesis with latent diffusion models, 2021.

\bibitem[Rombach et~al.(2022)Rombach, Blattmann, Lorenz, Esser, and Ommer]{rombach2022high}
Rombach, R., Blattmann, A., Lorenz, D., Esser, P., and Ommer, B.
\newblock High-resolution image synthesis with latent diffusion models.
\newblock In \emph{Proceedings of the IEEE/CVF conference on computer vision and pattern recognition}, pp.\  10684--10695, 2022.

\bibitem[Saharia et~al.(2022{\natexlab{a}})Saharia, Chan, Saxena, Li, Whang, Denton, Ghasemipour, Ayan, Mahdavi, Lopes, et~al.]{saharia2205photorealistic}
Saharia, C., Chan, W., Saxena, S., Li, L., Whang, J., Denton, E., Ghasemipour, S. K.~S., Ayan, B.~K., Mahdavi, S.~S., Lopes, R.~G., et~al.
\newblock Photorealistic text-to-image diffusion models with deep language understanding.
\newblock \emph{URL https://arxiv. org/abs/2205.11487}, 4, 2022{\natexlab{a}}.

\bibitem[Saharia et~al.(2022{\natexlab{b}})Saharia, Chan, Saxena, Li, Whang, Denton, Ghasemipour, Gontijo~Lopes, Karagol~Ayan, Salimans, et~al.]{saharia2022photorealistic}
Saharia, C., Chan, W., Saxena, S., Li, L., Whang, J., Denton, E.~L., Ghasemipour, K., Gontijo~Lopes, R., Karagol~Ayan, B., Salimans, T., et~al.
\newblock Photorealistic text-to-image diffusion models with deep language understanding.
\newblock \emph{Advances in Neural Information Processing Systems}, 35:\penalty0 36479--36494, 2022{\natexlab{b}}.

\bibitem[Salimans et~al.(2016)Salimans, Goodfellow, Zaremba, Cheung, Radford, and Chen]{salimans2016improved}
Salimans, T., Goodfellow, I., Zaremba, W., Cheung, V., Radford, A., and Chen, X.
\newblock Improved techniques for training gans.
\newblock \emph{Advances in neural information processing systems}, 29, 2016.

\bibitem[Touvron et~al.(2023)Touvron, Lavril, Izacard, Martinet, Lachaux, Lacroix, Rozi{\`e}re, Goyal, Hambro, Azhar, et~al.]{touvron2023llama}
Touvron, H., Lavril, T., Izacard, G., Martinet, X., Lachaux, M.-A., Lacroix, T., Rozi{\`e}re, B., Goyal, N., Hambro, E., Azhar, F., et~al.
\newblock Llama: Open and efficient foundation language models.
\newblock \emph{arXiv preprint arXiv:2302.13971}, 2023.

\bibitem[Wallace et~al.(2023)Wallace, Dang, Rafailov, Zhou, Lou, Purushwalkam, Ermon, Xiong, Joty, and Naik]{wallace2023diffusion}
Wallace, B., Dang, M., Rafailov, R., Zhou, L., Lou, A., Purushwalkam, S., Ermon, S., Xiong, C., Joty, S., and Naik, N.
\newblock Diffusion model alignment using direct preference optimization.
\newblock \emph{arXiv preprint arXiv:2311.12908}, 2023.

\bibitem[Wang et~al.(2022)Wang, Montoya, Munechika, Yang, Hoover, and Chau]{wangDiffusionDBLargescalePrompt2022}
Wang, Z.~J., Montoya, E., Munechika, D., Yang, H., Hoover, B., and Chau, D.~H.
\newblock {{DiffusionDB}}: {{A}} large-scale prompt gallery dataset for text-to-image generative models.
\newblock \emph{arXiv:2210.14896 [cs]}, 2022.
\newblock URL \url{https://arxiv.org/abs/2210.14896}.

\bibitem[Wei et~al.(2022)Wei, Wang, Schuurmans, Bosma, Xia, Chi, Le, Zhou, et~al.]{wei2022chain}
Wei, J., Wang, X., Schuurmans, D., Bosma, M., Xia, F., Chi, E., Le, Q.~V., Zhou, D., et~al.
\newblock Chain-of-thought prompting elicits reasoning in large language models.
\newblock \emph{Advances in Neural Information Processing Systems}, 35:\penalty0 24824--24837, 2022.

\bibitem[Wu et~al.(2023)Wu, Hao, Sun, Chen, Zhu, Zhao, and Li]{wu2023human}
Wu, X., Hao, Y., Sun, K., Chen, Y., Zhu, F., Zhao, R., and Li, H.
\newblock Human preference score v2: A solid benchmark for evaluating human preferences of text-to-image synthesis.
\newblock \emph{arXiv preprint arXiv:2306.09341}, 2023.

\bibitem[Xu et~al.(2023)Xu, Liu, Wu, Tong, Li, Ding, Tang, and Dong]{xu2023imagereward}
Xu, J., Liu, X., Wu, Y., Tong, Y., Li, Q., Ding, M., Tang, J., and Dong, Y.
\newblock Imagereward: Learning and evaluating human preferences for text-to-image generation, 2023.

\bibitem[You et~al.(2023)You, Li, Gu, Yin, Xue, and Dong]{you2023depicting}
You, Z., Li, Z., Gu, J., Yin, Z., Xue, T., and Dong, C.
\newblock Depicting beyond scores: Advancing image quality assessment through multi-modal language models.
\newblock \emph{arXiv preprint arXiv:2312.08962}, 2023.

\bibitem[Zhang et~al.(2023{\natexlab{a}})Zhang, Rao, and Agrawala]{zhang2023adding}
Zhang, L., Rao, A., and Agrawala, M.
\newblock Adding conditional control to text-to-image diffusion models.
\newblock In \emph{Proceedings of the IEEE/CVF International Conference on Computer Vision}, pp.\  3836--3847, 2023{\natexlab{a}}.

\bibitem[Zhang et~al.(2019)Zhang, Kishore, Wu, Weinberger, and Artzi]{zhang2019bertscore}
Zhang, T., Kishore, V., Wu, F., Weinberger, K.~Q., and Artzi, Y.
\newblock Bertscore: Evaluating text generation with bert.
\newblock \emph{arXiv preprint arXiv:1904.09675}, 2019.

\bibitem[Zhang et~al.(2023{\natexlab{b}})Zhang, Li, Sun, Liu, Min, and Zhai]{zhang2023perceptual}
Zhang, Z., Li, C., Sun, W., Liu, X., Min, X., and Zhai, G.
\newblock A perceptual quality assessment exploration for aigc images.
\newblock \emph{arXiv preprint arXiv:2303.12618}, 2023{\natexlab{b}}.

\bibitem[Zheng et~al.(2023)Zheng, Chiang, Sheng, Zhuang, Wu, Zhuang, Lin, Li, Li, Xing, Zhang, Gonzalez, and Stoica]{zheng2023judging}
Zheng, L., Chiang, W.-L., Sheng, Y., Zhuang, S., Wu, Z., Zhuang, Y., Lin, Z., Li, Z., Li, D., Xing, E.~P., Zhang, H., Gonzalez, J.~E., and Stoica, I.
\newblock Judging llm-as-a-judge with mt-bench and chatbot arena, 2023.

\bibitem[Zhu et~al.(2023)Zhu, Chen, Shen, Li, and Elhoseiny]{zhu2023minigpt}
Zhu, D., Chen, J., Shen, X., Li, X., and Elhoseiny, M.
\newblock Minigpt-4: Enhancing vision-language understanding with advanced large language models.
\newblock \emph{arXiv preprint arXiv:2304.10592}, 2023.

\end{thebibliography}
\bibliographystyle{icml2024}


\newpage
\appendix
\onecolumn
\section{Appendix}
\subsection{SynArtifact-1K Samples.}
\label{sec:SynArtifact-1K_Samples}
\begin{figure}[ht]
\centering
    \includegraphics[width=0.9\textwidth]{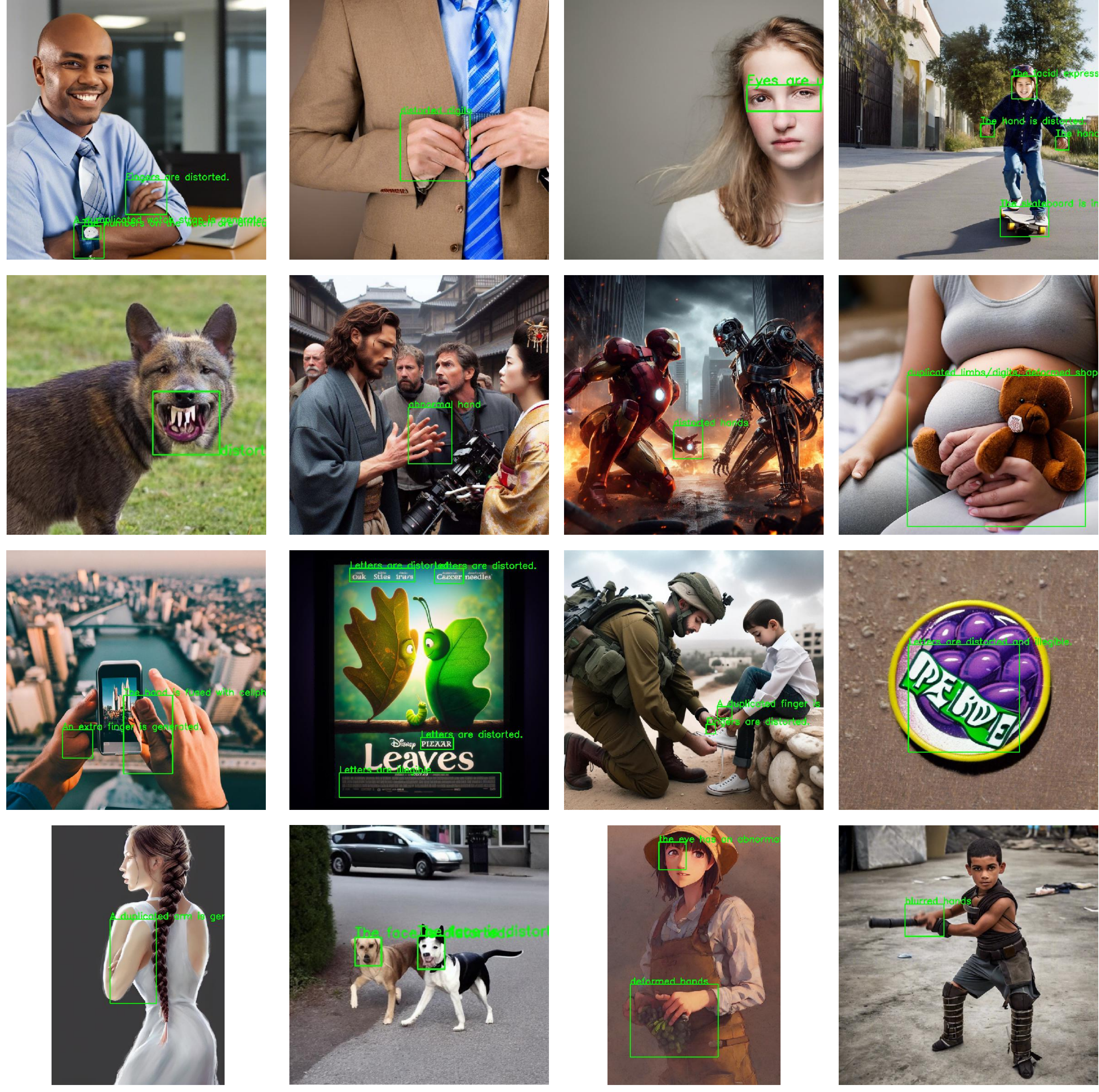}
\caption{\textbf{Visualization of Samples of SynArtifact-1K.} Each synthetic image with artifacts is annotated with bounding box and caption. Then, annotator maps the artifact caption into artifact taxonomy in \cref{fig:Artifact Taxonomy}.}
\label{fig:appendix_1}
\end{figure}

In our work, the main contribution is to propose SynArtifact-1K, the first synthetic image dataset including artifact labels, captions and coordinates. \cref{fig:appendix_1} shows examples of SynArtifact-1K. We can observe various kinds of artifacts in SynArtifact-1K, such as distorted hands, illegible letters and awkward facial expressions. Objects in SynArtifact-1K mainly contain human, animals and letters, which still remains challenging for generative models. Image styles in SynArtifact-1K are also various, such as art, realistic photos and movie poster. SynArtifact-1K can be utilized to classify and detect artifacts in synthetic images.       

\subsection{Instruction for Artifact Classification.}
\label{sec:Instruction_Artifact_Classification}
\begin{figure}[ht]
\centering
    \includegraphics[width=0.8\textwidth]{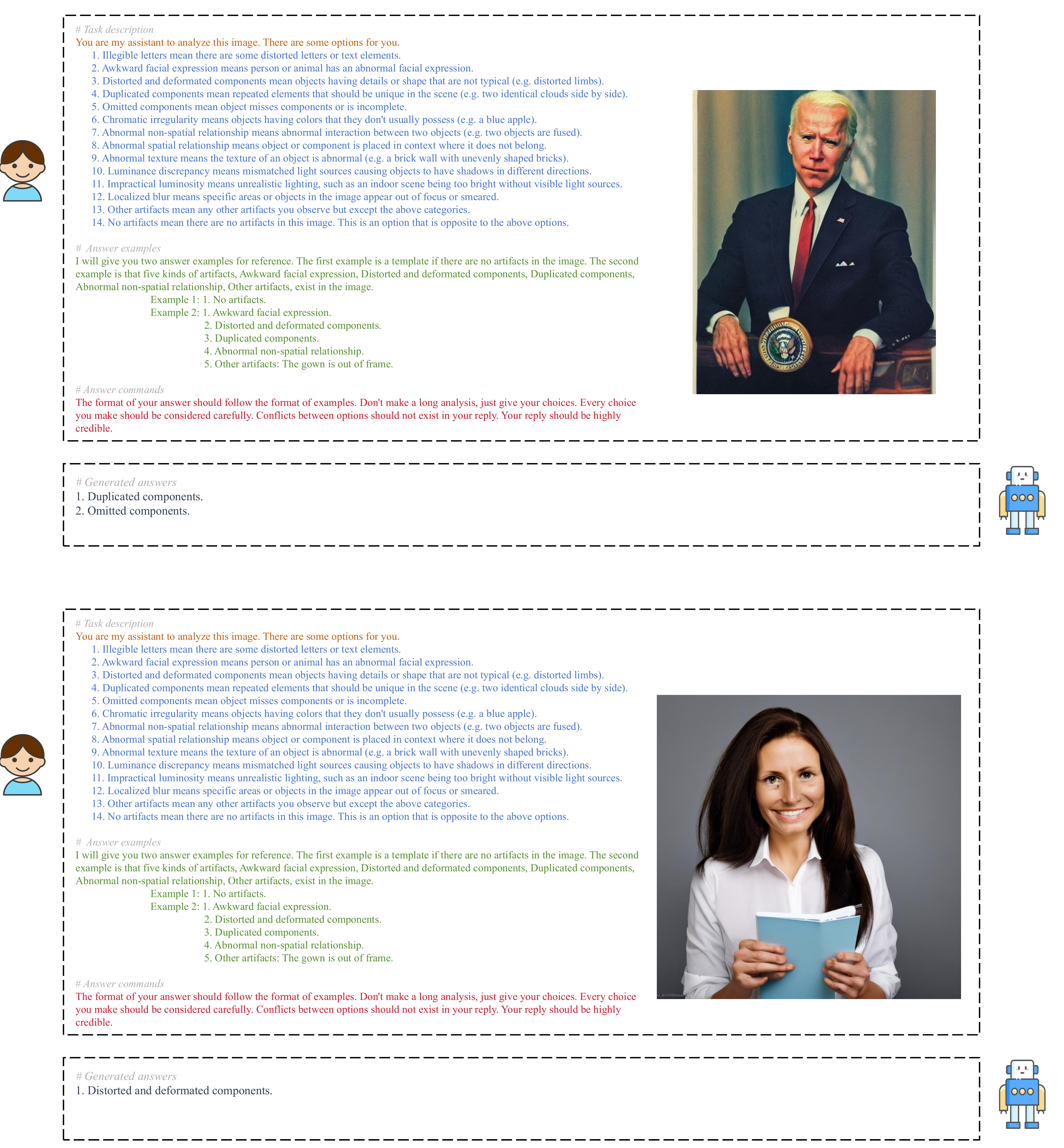}
\caption{\textbf{Instruction for Artifact Classification.} Instruction contains task description, all options, answer examples and answer commands. Answer only contains categories of artifacts in synthetic images.}
\label{fig:appendix_2}
\end{figure}

\cref{fig:appendix_2} shows complete instruction utilized for artifact classification. Instruction only contains one question-answer pair. We prompt Vision-Language Model with task description and all options. In order to standardize the format of answers, we provide two answers examples. One answer template is about no artifacts and another template is about synthetic image with artifacts. Furthermore, we also provide some answer commands to avoid in-context conflict within answer i.e. the answer includes both some artifact labels and no artifacts. 

\subsection{Instruction for Artifact Detection.}
\label{sec:appendix_Instruction_Artifact_Detection}
\begin{figure}[h]
\centering
    \includegraphics[width=0.9\textwidth]{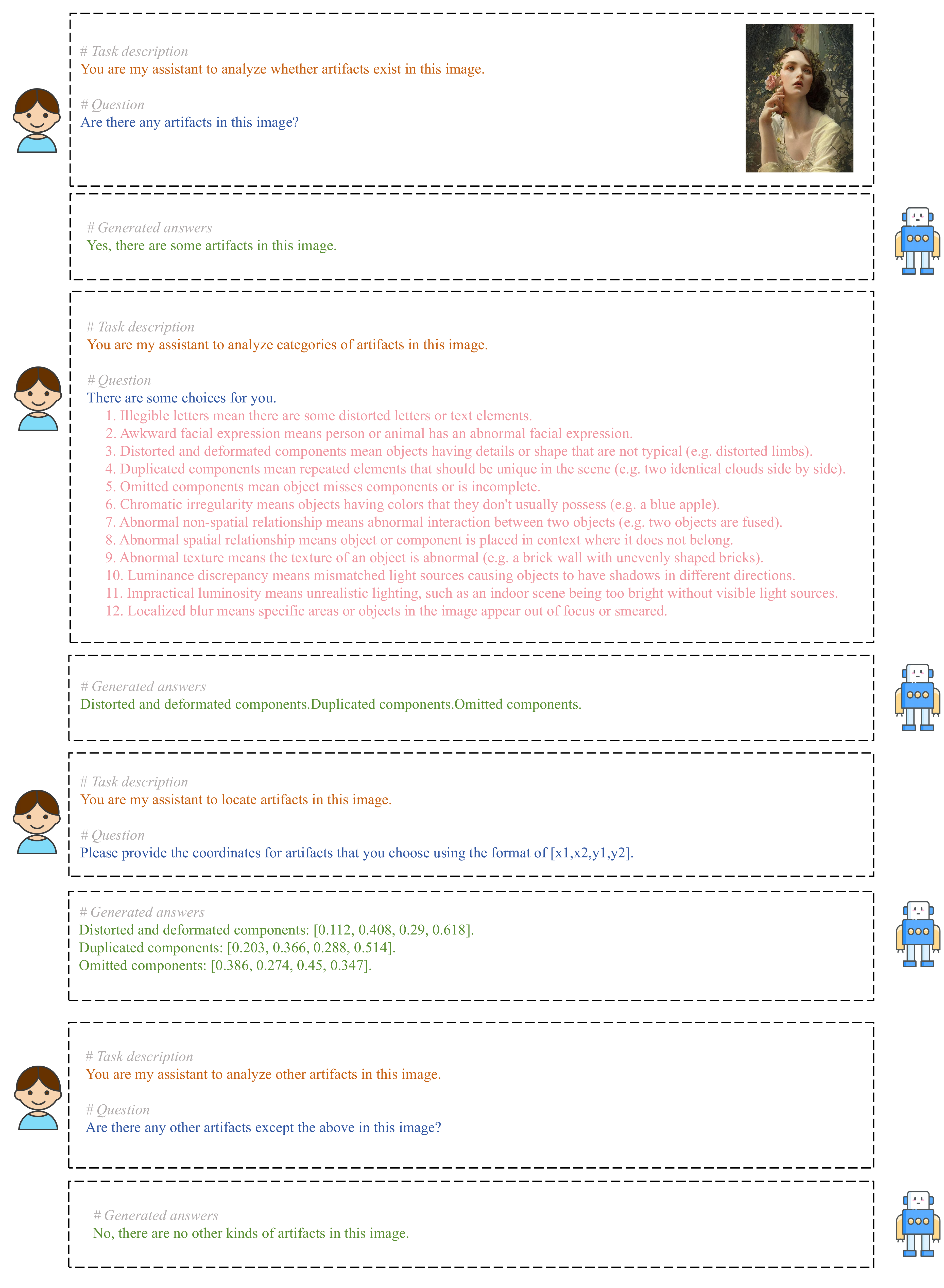}
\caption{\textbf{Instruction for Artifact Detection.} Instruction contains artifact judgement, artifact classification, artifact location and other artifact. Artifact classification question contains task description and all options. Locate artifact using the format of $[x_{1},y_{1},x_{2},y_{2}]$.}
\label{fig:appendix_3}
\end{figure}

\cref{fig:appendix_3} exhibits complete instruction for artifact detection. We divide artifact detection into four sub-tasks: artifact judgement, artifact classification, artifact location and other artifacts. Artifact judgement could be only answered utilizing \textit{``Yes"} or \textit{``No"}. Artifact classification should be answered utilizing various kinds of artifacts in \cref{fig:Artifact Taxonomy}. Furthermore, artifact location aims to locate artifacts that are selected in artifact classification using normalized coordinates $[x_{1},y_{1},x_{2},y_{2}]$. Considering the image process in CLIP will crop image to $336\times336$, we resize synthetic image into $336\times336$ with padding and normalize annotated coordinates to make instruction align with images. Finally, other artifact caption is also needed. To evaluate the performance of artifact detection, we load the pre-trained weights of LLaVA-v1.5-7B as the initialization and fine-tune it with a learning rate of 2e-5.

\clearpage
\subsection{Artifact Alleviated Samples.}
\begin{figure}[ht]
\centering
    \includegraphics[width=1.0\textwidth]{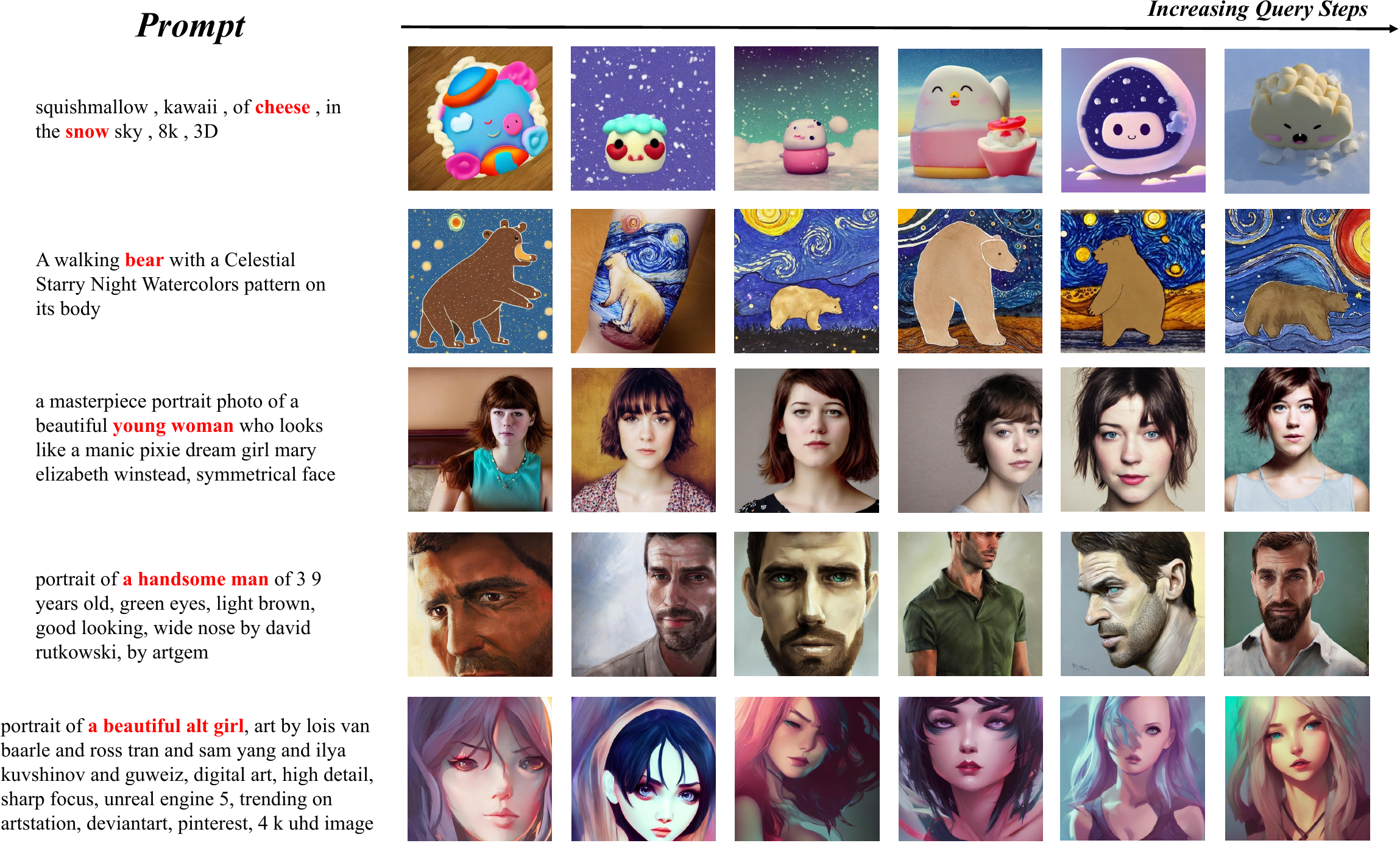}
\caption{\textbf{Visualization of Samples in Reinforcement Learning from Artifact Classification.} The leftmost image is produced with base Stable Diffusion v1.5. From left to right, it indicates images with an increasing number of query steps and artifact score.}
\label{fig:appendix_4}
\end{figure}

\cref{fig:appendix_4} shows samples with the process of fine-tuning diffusion model leveraging output of artifact classifier. We can observe that some artifacts have been alleviated. For example, bear's duplicated ears vanish (\textbf{rows 2}) and the man who is out of frame has been refined into the center of photo (\textbf{rows 4}).   

\end{document}